\documentclass[11pt]{article}

\usepackage[preprint]{acl}
\usepackage{xcolor}
\usepackage{hyperref}
\usepackage[most]{tcolorbox}
\definecolor{linkpink}{RGB}{236, 64, 122}
\definecolor{lightgraybg}{RGB}{238,238,238}
\usepackage{times}
\usepackage{latexsym}
\usepackage{float}
\usepackage[T1]{fontenc}

\usepackage[utf8]{inputenc}

\usepackage{microtype}

\usepackage{inconsolata}

\usepackage{graphicx}
\usepackage{svg}
\usepackage{booktabs}
\usepackage{multirow}
\usepackage{graphicx}
\usepackage[table]{xcolor}
\usepackage{cuted}
\usepackage{capt-of}
\usepackage{amsmath}
\usepackage{amssymb}

\newcommand{\method}{\textsc{Domino}}  
\newcommand{\tabsmall}{\scriptsize}
\newcommand{\secondarycolor}[1]{\textcolor{gray}{#1}}
\usepackage[table]{xcolor} 
\title{Domino: Decoupling Causal Modeling from Autoregressive Drafting in Speculative Decoding}

\author{
\begin{tabular}{c@{\hspace{1.2em}}c@{\hspace{1.2em}}c@{\hspace{1.2em}}c}
\textbf{Jianuo Huang}\textsuperscript{1,2*} &
\textbf{Yaojie Zhang}\textsuperscript{1,3*} &
\textbf{Qituan Zhang}\textsuperscript{4} \\
\textbf{Hao Lin}\textsuperscript{2} &
\textbf{Hanlin Xu}\textsuperscript{5} &
\textbf{Linfeng Zhang}\textsuperscript{1\textdagger}
\end{tabular}
\\[0.5em]
\textsuperscript{1}EPIC Lab, Shanghai Jiao Tong University \quad
\textsuperscript{2}School of Software Engineering, HUST \\
\textsuperscript{3}UESTC
\quad
\textsuperscript{4}Fudan University \quad
\textsuperscript{5}Huawei
\\[0.3em]
\textsuperscript{*}Equal contribution. \quad
\textsuperscript{\textdagger}Corresponding author.
}

\begin{document}
\maketitle


\begin{abstract}

Speculative decoding accelerates LLM inference by drafting multiple tokens and verifying them in parallel with the target model. However, its practical speedup is constrained by the trade-off between draft quality and drafting cost: autoregressive drafters model causal dependencies among draft tokens but incur sequential overhead, while parallel drafters reduce drafting cost but weaken intra-block dependency modeling. In this paper, we propose Domino, a speculative decoding framework that decouples causal dependency modeling from expensive autoregressive draft execution. Domino first uses a parallel draft backbone to produce preliminary draft distributions for the entire block, and then applies a lightweight Domino head to refine them with prefix-dependent causal information. To stabilize teacher-forced causal encoding, we further introduce a base-anchored training curriculum that first strengthens the parallel backbone and then gradually shifts optimization toward the causally corrected final distribution. Experiments on Qwen3 models show that Domino achieves up to \(5.49\times\) end-to-end speedup under the Transformers backend and up to \(5.8\times\) throughput speedup under SGLang serving. 
\begin{center}
\vspace{-0.5em}
{\small
\textbf{Links:} \ 
\href{https://github.com/jianuo-huang/Domino}{\textcolor{linkpink}{Code}}
\ (GitHub) \ \textnormal{\textbar }\ 
\href{https://huggingface.co/collections/Huang2020/domino}{\textcolor{linkpink}{Models}}
\ \textnormal{(Hugging Face)}
}
\vspace{-0.5em}
\end{center}
\end{abstract}
\vspace{-4mm}
\section{Introduction}
While large language models have achieved expert-level performance on reasoning, coding, and long-context tasks \citep{singh2025openai}, their standard 
autoregressive decoding remains inherently sequential. This process is often memory-bound, leaving the massive parallelism of modern GPUs underutilized and leading to 
high inference latency.
\begin{figure}[t]
    \centering
    \includegraphics[width=\columnwidth]{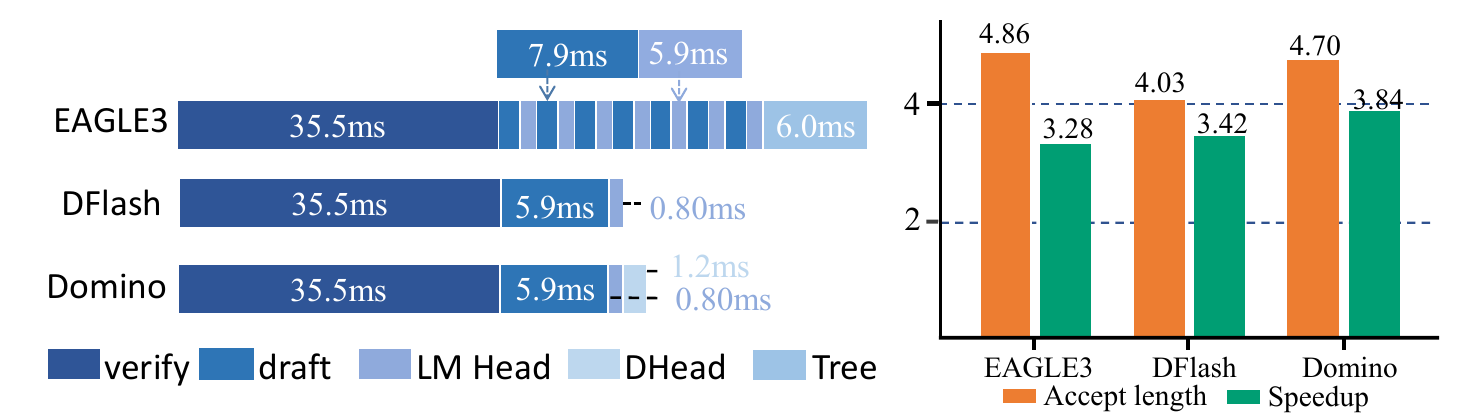}
    \caption{
    Latency breakdown and performance comparison on Qwen3-8B under a 16-token speculative decoding budget.
    Left: per-step latency breakdown measured on an A100 GPU with context length 1024, where \emph{Verify} denotes the target-model verification latency and \emph{Draft} denotes the draft-model forward latency.
    \emph{LM Head}, \emph{DHead}, and \emph{Tree} denote the output projection, Domino head, and tree construction/sampling overheads, respectively.
    Right: acceptance length and end-to-end speedup evaluated on GSM8K.
    All three draft models are trained on the same dataset.
    }

    \label{fig:draft_overhead}
    \vspace{-2mm}
\end{figure}
\begin{figure*}[t]
  \centering
  \includegraphics[width=1.0\linewidth]{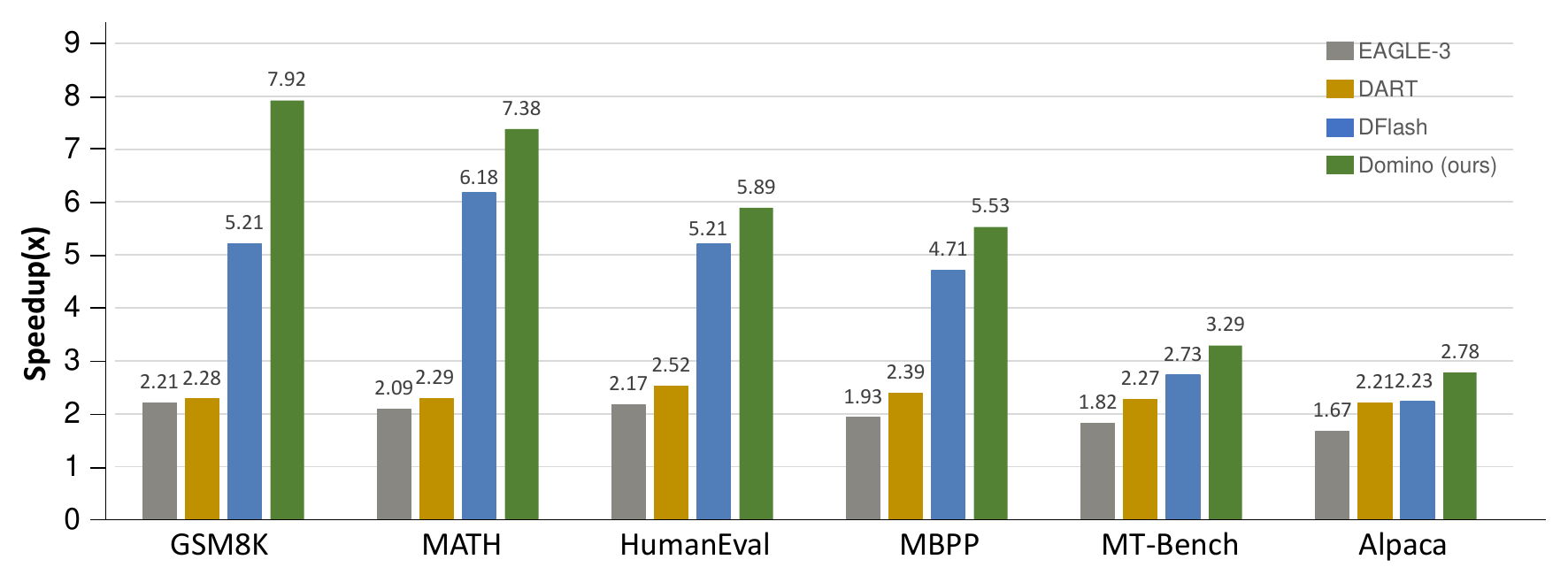}
  \caption{Speedup comparison of Domino, DFlash, and EAGLE-3 relative to autoregressive decoding on Qwen3-8B using the Transformers backend.}
  \label{fig:domino_intro}
\end{figure*}
To alleviate this bottleneck, speculative decoding has emerged as a widely adopted strategy for accelerating LLM inference \citep{leviathan2023fast}. It leverages a lightweight draft model to propose multiple future tokens, which are then verified in parallel by the target model within a single forward pass. By reducing the number of expensive invocations of the target model, this draft-then-verify mechanism preserves the target model's output distribution while improving throughput. Subsequent work has explored various drafting strategies, including head-based multi-token prediction \citep{cai2024medusa} and autoregressive draft models \citep{li2024eagle, li2024eagle2, li2025eagle3}. Across these methods, the resulting speedup is jointly determined by two factors: \emph{draft quality and drafting cost}. Higher-quality drafts yield longer acceptance lengths, but the drafting process itself introduces additional overhead that can diminish or even negate these gains.

This quality--cost trade-off is especially evident in autoregressive drafting methods such as the EAGLE series \citep{li2024eagle, li2024eagle2, li2025eagle3}. By generating draft tokens sequentially, autoregressive drafters explicitly model causal dependencies within the draft block, improving alignment with the target model's autoregressive distribution and yielding long acceptance lengths. However, generating \(k\) draft tokens requires \(k\) sequential draft steps, each involving a draft-model forward pass and a full-vocabulary LM-head projection. This cost grows linearly with draft length and can offset the gains from higher acceptance length, limiting the speedup obtained by scaling draft length or draft capacity \citep{zhao2025fr,yan2025scaling}.


Existing methods address this issue from different directions. FR-Spec \citep{zhao2025fr} and SpecVocab \citep{williams2026speculative} reduce the cost of full-vocabulary projection through static or dynamic vocabulary selection, but still retain autoregressive draft execution. In contrast, DFlash \citep{chen2026dflash} fully parallelizes drafting by producing an entire draft block in one forward pass, avoiding repeated per-token draft-model and LM-head calls. However, removing sequential draft dependencies weakens intra-block causal modeling and can reduce the alignment between the draft distribution and the target model distribution.

Figure~\ref{fig:draft_overhead} illustrates this trade-off. 
EAGLE-3 obtains a high acceptance length of \(4.86\), but its sequential draft execution and tree construction limit the speedup to \(3.28\times\). 
DFlash reduces drafting overhead through block-parallel generation and improves speedup to \(3.42\times\), but its acceptance length decreases to \(4.03\). 
These results suggest that causal dependency modeling is useful for draft quality, but the standard autoregressive implementation makes it expensive. 
This raises a natural question: can we retain the draft-quality benefit of causal dependency modeling while preserving the low drafting cost of block-parallel generation?

To answer this question, we propose Domino, a lightweight causal correction framework that decouples causal dependency modeling from expensive autoregressive draft execution. Instead of generating draft tokens sequentially, Domino keeps the main drafting computation parallel: a parallel draft backbone first produces preliminary draft distributions for the entire block. On top of these distributions, Domino applies a lightweight Domino head to inject causal information. The Domino head uses a causal encoder to summarize previously drafted tokens and a low-rank correction head to refine the draft distributions through residual correction, avoiding repeated draft-model execution and another expensive full LM-head computation.

This design allows Domino to recover useful intra-block causal dependency while preserving the efficiency of block-parallel drafting. As shown in Figure~\ref{fig:draft_overhead}, compared with DFlash, Domino adds only 56M parameters (+5.3\%) and incurs only a 2.8\% increase in total draft-then-verify latency. Simultaneously, it improves average acceptance length by 16.6\% and end-to-end speedup by 12.3\%. 

Figure~\ref{fig:domino_intro} provides a preview of the empirical gains. 
On representative math, code, and chat benchmarks, Domino consistently outperforms EAGLE-3, DART, and DFlash, achieving up to \(7.92\times\) speedup on GSM8K and improving over DFlash from \(5.21\times\) to \(7.92\times\). 
These results suggest that lightweight causal correction improves draft quality while preserving the efficiency of block-parallel drafting.



\section{Related Work}
\noindent\textbf{Speculative Decoding.}
Speculative decoding accelerates autoregressive LLM inference by using a draft model to propose candidate tokens and a target model to verify them in parallel. Early approaches use a smaller language model as the drafter, which generates candidate tokens autoregressively before verification by the target model \citep{leviathan2023fast, chen2023accelerating}. Subsequent methods improve this basic draft-then-verify pipeline through tree-based verification, better serving systems, and more efficient draft model designs \citep{miao2023specinfer, cai2024medusa, li2024eagle}. These works establish the general framework of speculative decoding, where the final speedup depends on both the acceptance length and the cost of generating draft tokens.

\noindent\textbf{Autoregressive and Efficient Drafting.}
A representative line of speculative decoding methods improves draft quality through autoregressive drafting. The EAGLE series generates draft tokens sequentially, allowing each token to depend on previous draft tokens and better match the target model's autoregressive distribution \citep{li2024eagle, li2024eagle2, li2025eagle3}. Other methods reduce drafting overhead from different angles: Medusa uses lightweight parallel decoding heads \citep{cai2024medusa}, Hydra introduces sequentially-dependent heads to inject causal information into head-based drafting \citep{ankner2024hydra}, and FR-Spec and SpecVocab reduce full-vocabulary projection costs through static or dynamic vocabulary selection \citep{zhao2025fr,williams2026speculative}. 

\noindent\textbf{Parallel and Non-Autoregressive Drafting.}
Speculative Diffusion Decoding first explores discrete diffusion models as parallel drafters for speculative decoding \citep{christopher2025speculative}. DiffuSpec further uses pretrained diffusion language models as training-free drafters \citep{li2025diffuspec}. PARD adapts autoregressive models into parallel draft models, allowing multiple future tokens to be predicted in a single draft forward pass \citep{an2026pard}.  More recently, DART predicts token distributions for multiple future positions in parallel and uses  tree pruning to construct draft candidates \citep{liu2026dart}. DFlash adopts a block-diffusion drafter that produces an entire draft block in a single forward pass, avoiding repeated calls to both the draft model and the full LM head \citep{chen2026dflash}. These methods substantially reduce drafting overhead, but fully parallel drafting weakens intra-block causal dependencies, making it harder to match the target model's autoregressive distribution. In contrast, Domino keeps the main draft computation parallel while reintroducing causal information through a lightweight correction branch.
\section{Preliminaries}
\subsection{Speculative Decoding and Speedup}
Speculative decoding accelerates autoregressive inference by using a draft model \(M_d\) to propose multiple future tokens, which are then verified in parallel by the target model \(M_t\). At each decoding cycle, the draft model proposes \(\gamma\) candidate tokens. The target model evaluates these candidates in a single forward pass and accepts the longest valid prefix according to the standard speculative verification rule. We denote by \(\tau \in [1, \gamma+1]\) the expected number of tokens advanced per cycle, including the bonus token produced by the target model.

The average per-token latency of speculative decoding can be written as
\[
L_{\mathrm{spec}}
=
\frac{T_{\mathrm{draft}} + T_{\mathrm{verify}}}{\tau},
\]
where \(T_{\mathrm{draft}}\) is the time spent generating draft tokens, and \(T_{\mathrm{verify}}\) is the time spent verifying them with the target model. Let \(L_{\mathrm{target}}\) denote the per-token latency of standard autoregressive decoding with the target model. The resulting speedup is
\[
\eta
=
\frac{L_{\mathrm{target}}}{L_{\mathrm{spec}}}
=
\frac{\tau L_{\mathrm{target}}}
{T_{\mathrm{draft}} + T_{\mathrm{verify}}}.
\]

This expression highlights two key factors that determine the final speedup. First, increasing the acceptance length \(\tau\) allows each target-model invocation to advance more tokens. Second, reducing the draft cost \(T_{\mathrm{draft}}\) prevents the drafting stage from offsetting the benefit of parallel verification. Therefore, an effective drafter must be both accurate enough to achieve long acceptance lengths and efficient enough to keep drafting overhead low.

\subsection{Autoregressive Drafting}

Autoregressive drafters generate candidate tokens sequentially. Given a prefix \(x_{\le t}\), an autoregressive drafter factorizes the draft distribution as
\[
q_{\mathrm{AR}}(x_{t+1:t+\gamma} \mid x_{\le t})
=
\prod_{i=1}^{\gamma}
q(x_{t+i} \mid x_{<t+i}),
\]
where \(x_{<t+i}\) includes the current prefix \(x_{\le t}\) and previously drafted tokens.
This factorization mirrors the target model's autoregressive prediction process, where each token is conditioned on all previous tokens. As a result, autoregressive drafting can explicitly use previously drafted tokens when predicting later draft positions, leading to higher draft quality and longer acceptance length.

However, this modeling advantage comes with a sequential execution cost. More concretely, for an autoregressive drafter, generating \(\gamma\) draft tokens requires \(\gamma\) draft steps, each consisting of a draft-model forward computation followed by an LM-head projection. Denoting the average latency of these two operations by \(t_{\mathrm{net}}\) and \(t_{\mathrm{head}}\), respectively, the total drafting cost can be approximated as
\[
T_{\mathrm{draft}}^{\mathrm{AR}}
\approx
\gamma \cdot
\left(
t_{\mathrm{net}} + t_{\mathrm{head}}
\right).
\]
This cost grows approximately linearly with the speculation budget \(\gamma\), and can be further amplified when the draft model becomes deeper or the vocabulary size is large \citep{zhao2025fr,yan2025scaling}. Therefore, although autoregressive drafters often achieve high acceptance lengths, their repeated executions can offset the gain from longer accepted prefixes and limit the achievable speedup.

\subsection{Parallel Drafting}

Unlike autoregressive drafters that factorize the draft distribution from left to right, a parallel drafter \citep{chen2026dflash,liu2026dart} directly predicts the block-level conditional distribution
\[
q_{\mathrm{PAR}}(x_{t+1:t+\gamma} \mid x_{\le t}),
\]
thereby generating multiple draft tokens in parallel. Its drafting cost can be approximated as
\[
T_{\mathrm{draft}}^{\mathrm{PAR}}
\approx
t_{\mathrm{net}}^{\mathrm{block}} + t_{\mathrm{head}}^{\mathrm{block}},
\]
where \(t_{\mathrm{net}}^{\mathrm{block}}\) and \(t_{\mathrm{head}}^{\mathrm{block}}\) denote the average latency of the block-level draft-model forward pass and the corresponding LM-head projection, respectively. Unlike autoregressive drafting, they are incurred once for the whole draft block rather than repeated \(\gamma\) times, enabling better GPU utilization through parallel computation.

The verification stage remains similar for both autoregressive and parallel drafters: the target model evaluates the proposed draft block in parallel and accepts the longest valid prefix. The main difference lies in draft generation. Autoregressive drafters usually achieve longer acceptance lengths because they explicitly model causal dependencies within the draft sequence, but their draft cost grows with the number of generated tokens. Parallel drafters reduce or eliminate this sequential drafting cost, but often weaken intra-block causal dependencies among draft tokens. Consequently, they may require larger draft capacity to reach acceptance length comparable to autoregressive drafters \citep{chen2026dflash}.

\section{Methodology}

\begin{figure*}[t]
  \centering
  \includegraphics[width=1.0\linewidth]{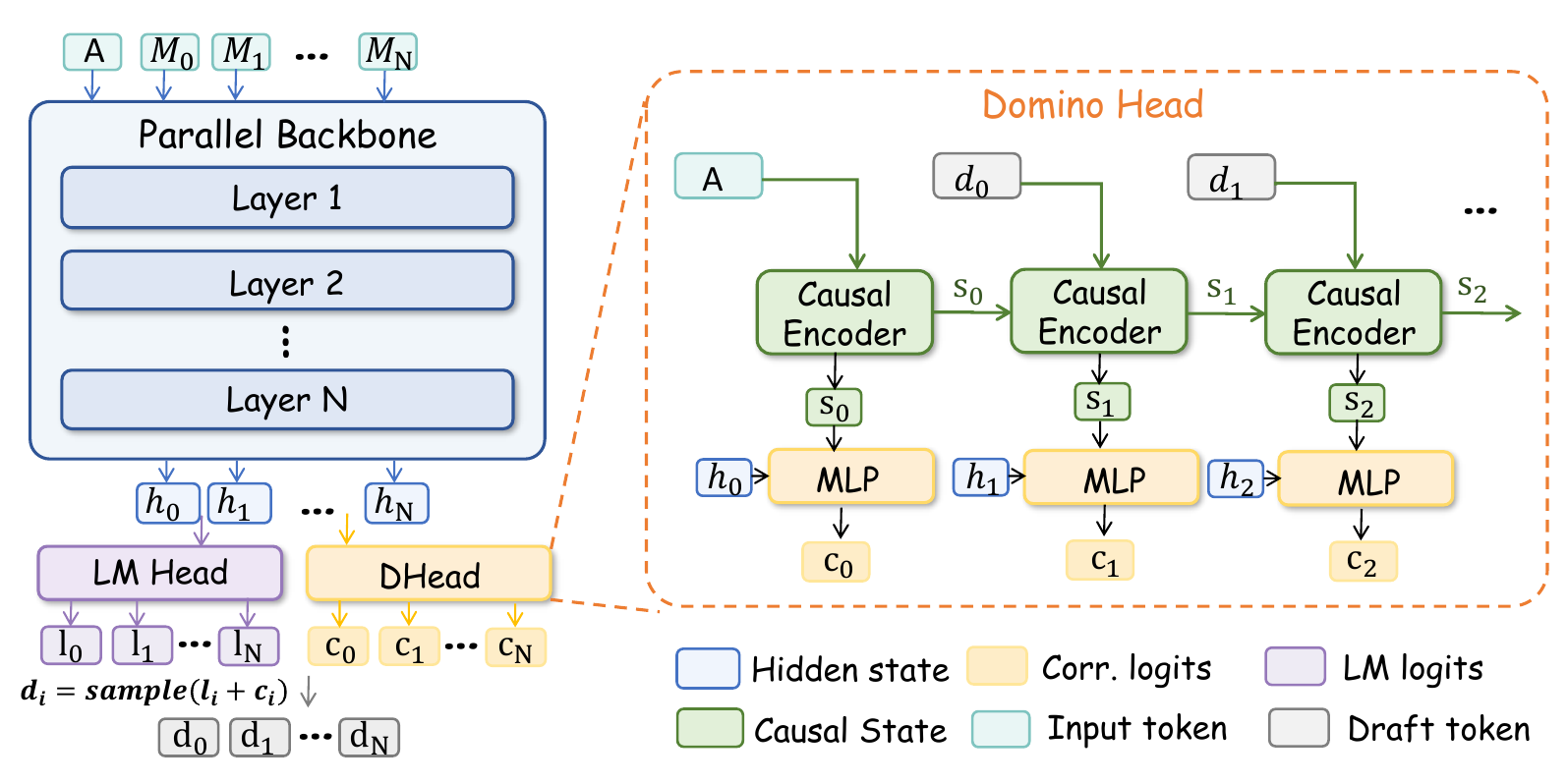}
    \caption{
    Overview of Domino. 
    The parallel backbone produces hidden states for the whole draft block in one forward pass. 
    The Domino head sequentially updates a causal state from previously sampled draft tokens and generates correction logits \(c_i\), which refine the base logits \(l_i\). 
    Each draft token is sampled from the final logits \(l_i+c_i\).
    }
  \label{fig:pipe_figure}
\end{figure*}

\subsection{Architecture of Domino}
\label{sec:model_structure}
Figure~\ref{fig:pipe_figure} gives an overview of Domino. Our method contains two components: a parallel draft backbone and a \emph{Domino head}. The parallel draft backbone generates preliminary distributions for all positions in the draft block in a single parallel computation. The Domino head then refines these preliminary distributions by propagating causal information across draft positions.

\subsubsection{Parallel Draft Backbone}
We instantiate the parallel draft backbone with the DFlash architecture \citep{chen2026dflash}, which generates block-level draft representations from target context features and masked block inputs. Given a verified prefix \(x_{\le t}\), we use the last verified token \(x_t\) as the anchor and construct a masked draft block
\[
\tilde{x}_{t:t+B-1}
=
[x_t, \texttt{[MASK]}, \ldots, \texttt{[MASK]}],
\]
where the remaining \(B-1\) positions correspond to future draft tokens. The backbone takes the target-model context features \(C_t\) extracted from the verified prefix and the embeddings of the masked draft block as input, and produces block-level hidden states in one non-autoregressive forward pass:
\[
H_{t:t+B-1}
=
\mathrm{Backbone}
\left(
C_t,\,
\mathrm{Embed}(\tilde{x}_{t:t+B-1})
\right).
\]
The preliminary logits for future positions are computed by applying the frozen target LM head:
\[
L_i^{\mathrm{base}} = \mathrm{LMHead}(H_i),
\quad i=t+1,\ldots,t+B-1,
\]
where \(\mathrm{LMHead}\) denotes the LM head of the target model.
The parallel backbone produces hidden representations for the entire draft block in parallel, from which base logits are computed using the target LM head.

\subsubsection{Domino Head}

The Domino head injects causal information into the parallel base logits. It contains a causal encoder and a low-rank correction head.

\paragraph{Causal Encoder.}
For each draft position \(i\), we use a lightweight GRU to summarize the embeddings of preceding draft tokens \citep{cho2014learning}. Let \(E_j\) denote the token embedding at position \(j\). The causal state before predicting position \(i\) is
\[
S_{i-1} = \mathrm{GRU}(E_{\le i-1}),
\]
where \(S_{i-1} \in \mathbb{R}^{d_s}\) represents the prefix-dependent information available to position \(i\). In practice, the GRU is lightweight; in our implementation, we use a hidden dimension of 1024. This causal state allows later draft positions to receive information from earlier draft tokens without invoking a full autoregressive draft model.

\paragraph{Low-Rank Correction Head.}
Given the base representation \(H_i\) and causal state \(S_{i-1}\), the Domino head produces a logit-space residual correction through a low-rank bottleneck:
\[
\Delta L_i
=
W_2 \, \sigma\!\left(W_1 [H_i; S_{i-1}]\right),
\]
where \(W_1\) projects the concatenated representation into a low-rank hidden space of dimension \(r\), \(W_2\) maps the low-rank representation to the vocabulary space, and \(\sigma\) is the SiLU activation. The final draft logits are then computed as
\[
L_i = L_i^{\mathrm{base}} + \Delta L_i.
\]
In our implementation, \(r=256\). Since the correction is computed from a low-rank hidden space, it is much cheaper than repeatedly applying a full LM head in an autoregressive draft loop.

We apply correction in logit space rather than hidden space. A hidden-space correction would require applying the full LM head again after each causal update, reintroducing the expensive full-head computation into the sequential branch. In contrast, our logit-space correction keeps the base LM-head computation parallel and restricts the causal branch to a low-rank residual update.

\subsection{Training}
\label{sec:training}
We describe two training choices for Domino, each addressing a different failure mode.
\paragraph{Teacher-Forced Causal Encoding.}
The causal encoder consumes the preceding draft tokens within the current block. A natural choice is to feed it self-generated prefixes during training, following the training-time testing (TTT) strategy used in EAGLE-3 \citep{li2025eagle3}, which simulates multi-step draft generation during training. Instead, we use teacher forcing and feed the encoder ground-truth token embeddings.

There are two reasons for this choice. First, self-generated prefixes can be noisy and often incorrect, especially early in training. Supervising the model to map such corrupted prefixes to the ground-truth next token creates an input--output mapping that does not exist in the underlying data distribution \citep{huszar2015not}. This mismatch can degrade the causal representations learned by the GRU.

Second, teacher forcing is better aligned with the acceptance mechanism of speculative decoding. A draft token at position \(i\) contributes to the acceptance length only when all preceding draft tokens have already been verified as correct. Therefore, the correction at position \(i\) only matters in the accepted-prefix regime, where the preceding draft tokens match the target sequence. Training the causal encoder on ground-truth prefixes directly focuses learning on this regime; corrections conditioned on incorrect prefixes are less relevant, since those positions will be rejected by verification. We compare teacher forcing with TTT in Section~\ref{sec: abl_training}.

\paragraph{Base-anchored curriculum.}
Teacher forcing introduces another failure mode. Since the correction branch receives clean prefixes during training, directly optimizing only the final-logit loss can allow the correction branch to shortcut the parallel backbone. In this case, the base logits may become weak, and the final prediction may rely too heavily on the correction branch. As shown in Figure~\ref{fig:training_strategy}, this leads to a collapse of the parallel backbone loss, while the proposed base-anchored curriculum keeps the backbone loss decreasing steadily.

To prevent this collapse, we jointly supervise the base and final logits with a time-varying weight:
  \[
  \mathcal{L}
  =
  (1-\lambda_t)\,\mathcal{L}_{\mathrm{final}}
  +
  \lambda_t\,\mathcal{L}_{\mathrm{base}},
  \]
where \(\mathcal{L}_{\mathrm{final}}\) and \(\mathcal{L}_{\mathrm{base}}\) are cross-entropy losses computed from \(L_i\) and \(L_i^{\mathrm{base}}\), respectively. We linearly anneal \(\lambda_t\) from
\(1\) to \(0\) over the course of training, so the objective is initially anchored on the base logits---forcing the parallel backbone to learn a strong base distribution---and gradually shifts to the final
logits as the Domino head takes over the residual correction. We ablate this curriculum against direct final-loss training in Section~\ref{sec: abl_training}.

Following common practice in block-level speculative decoding \citep{chen2026dflash}, both \(\mathcal{L}_{\mathrm{base}}\) and \(\mathcal{L}_{\mathrm{final}}\) use a position-wise exponential
decay \(w_k = \exp(-k/\gamma)\) to prioritize earlier draft positions, whose acceptance gates the rest of the block.
  
\subsection{Efficient Runtime Implementation}
\label{sec:runtime_optimization}

To minimize the overhead of the Domino head, we implement its correction loop with fused Triton kernels and CUDA Graphs. This reduces kernel-launch and Python-level overhead during rollout. Under the latency setting in Figure~\ref{fig:draft_overhead}, the Domino-head latency decreases from \(2.64\)ms to \(1.20\)ms.

\begin{table*}[t]
    \caption{
    Decoding speedup over vanilla autoregressive decoding and average acceptance length ($\tau$) on Qwen3 models with a maximum of 2048 generated tokens. 
    Parenthesized values indicate the draft tree size for EAGLE-3 and DART, and the draft block size for DFlash and \method. 
    The average is computed over all listed benchmarks. 
    For EAGLE-3, DFlash, and DART, we use the publicly released checkpoints listed in Table~\ref{tab:baseline_checkpoints}.
    }
    \label{tab:main-results}
    \resizebox{\linewidth}{!}{
        \scriptsize
        \centering
        \setlength{\tabcolsep}{1.2pt}
        \begin{tabular}{c l @{\hspace{1.0em}} cc cc cc @{\hspace{1.0em}} cc cc cc @{\hspace{1.0em}} cc cc @{\hspace{1.0em}} cc}
            \toprule
            \multirow{2}[2]{*}{Model} & \multirow{2}[2]{*}{\quad Method} 
            & \multicolumn{6}{c@{\hspace{1.0em}}}{\sc{Math}} 
            & \multicolumn{6}{c@{\hspace{1.0em}}}{\sc{Code}} 
            & \multicolumn{4}{c@{\hspace{1.0em}}}{\sc{Chat}}
            & \multicolumn{2}{c}{\sc{Overall}} \\
            \cmidrule(lr){3-8}
            \cmidrule(lr){9-14}
            \cmidrule(lr){15-18}
            \cmidrule(lr){19-20}
            & & \multicolumn{2}{c}{GSM8K} 
            & \multicolumn{2}{c}{MATH-500} 
            & \multicolumn{2}{c@{\hspace{1.0em}}}{AIME25} 
            & \multicolumn{2}{c}{HumanEval} 
            & \multicolumn{2}{c}{MBPP} 
            & \multicolumn{2}{c@{\hspace{1.0em}}}{LCB} 
            & \multicolumn{2}{c}{MT-Bench}
            & \multicolumn{2}{c@{\hspace{1.0em}}}{Alpaca}
            & \multicolumn{2}{c}{\textit{Avg.}} \\
            \midrule

            \multicolumn{2}{c}{Temperature = 0} 
            & \tabsmall Speedup & \tabsmall $\tau$ 
            & \tabsmall Speedup & \tabsmall $\tau$ 
            & \tabsmall Speedup & \tabsmall $\tau$ 
            & \tabsmall Speedup & \tabsmall $\tau$ 
            & \tabsmall Speedup & \tabsmall $\tau$ 
            & \tabsmall Speedup & \tabsmall $\tau$ 
            & \tabsmall Speedup & \tabsmall $\tau$
            & \tabsmall Speedup & \tabsmall $\tau$
            & \tabsmall Speedup & \tabsmall $\tau$  \\
            
            \midrule
            \multirow{5}{*}{Qwen3-4B}
            & \mbox{EAGLE-3~\tabsmall\secondarycolor{(16)}}
            & 2.24$\times$ & 3.32 
            & 2.10$\times$ & 3.11 
            & 2.08$\times$ & 3.10 
            & 2.09$\times$ & 3.09 
            & 2.02$\times$ & 2.99 
            & 1.95$\times$ & 2.93 
            & 1.95$\times$ & 2.94 
            & 1.87$\times$ & 2.83
            & 2.04$\times$ & 3.04 \\

            & \mbox{EAGLE-3~\tabsmall\secondarycolor{(60)}} 
            & 2.57$\times$ & 3.83 
            & 2.40$\times$ & 3.59 
            & 2.36$\times$ & 3.53 
            & 2.36$\times$ & 3.53 
            & 2.30$\times$ & 3.44 
            & 2.19$\times$ & 3.31 
            & 2.25$\times$ & 3.41 
            & 2.14$\times$ & 3.29
            & 2.32$\times$ & 3.49 \\

            & \mbox{DART~\tabsmall\secondarycolor{(60)}}
            & 2.16$\times$ & 2.65 
            & 2.20$\times$ & 2.63 
            & 2.13$\times$ & 2.59 
            & 2.48$\times$ & 2.99 
            & 2.47$\times$ & 3.01 
            & 2.23$\times$ & 2.76 
            & 2.19$\times$ & 2.69 
            & 2.16$\times$ & 2.78
            & 2.25$\times$ & 2.76 \\

            & \mbox{DFlash~\tabsmall\secondarycolor{(16)}}
            & 5.18$\times$ & 6.51 
            & 6.10$\times$ & 7.79 
            & 5.87$\times$ & 7.47 
            & 5.23$\times$ & 6.63 
            & 4.77$\times$ & 6.12 
            & \textbf{5.48$\times$} & \textbf{7.15} 
            & 2.72$\times$ & 4.12 
            & 2.21$\times$ & 3.12
            & 4.70$\times$ & 6.11 \\

            & \mbox{\method~\tabsmall\secondarycolor{(16)}}
            & \textbf{8.02$\times$} & \textbf{10.12} 
            & \textbf{7.14$\times$} & \textbf{9.01} 
            & \textbf{5.97$\times$} & \textbf{7.49} 
            & \textbf{5.66$\times$} & \textbf{7.02} 
            & \textbf{5.59$\times$} & \textbf{7.11} 
            & 5.33$\times$ & 6.86 
            & \textbf{3.28$\times$} & \textbf{5.12} 
            & \textbf{2.80$\times$} & \textbf{3.88}
            & \textbf{5.47$\times$} & \textbf{7.08} \\

            \midrule
            \multirow{5}{*}{Qwen3-8B}
            & \mbox{EAGLE-3~\tabsmall\secondarycolor{(16)}}
            & 2.21$\times$ & 3.27 
            & 2.09$\times$ & 3.10 
            & 2.07$\times$ & 3.07 
            & 2.17$\times$ & 3.21 
            & 1.93$\times$ & 2.86 
            & 1.80$\times$ & 2.95 
            & 1.82$\times$ & 2.75 
            & 1.67$\times$ & 2.53
            & 1.97$\times$ & 2.97 \\

            & \mbox{EAGLE-3~\tabsmall\secondarycolor{(60)}} 
            & 2.56$\times$ & 3.80 
            & 2.42$\times$ & 3.61 
            & 2.41$\times$ & 3.59 
            & 2.50$\times$ & 3.74 
            & 2.22$\times$ & 3.31 
            & 2.03$\times$ & 3.12 
            & 2.07$\times$ & 3.17 
            & 1.88$\times$ & 2.90
            & 2.26$\times$ & 3.41 \\

            & \mbox{DART~\tabsmall\secondarycolor{(60)}}
            & 2.28$\times$ & 2.71 
            & 2.29$\times$ & 2.70 
            & 2.11$\times$ & 2.59 
            & 2.52$\times$ & 2.95 
            & 2.39$\times$ & 2.98 
            & 2.24$\times$ & 2.78 
            & 2.27$\times$ & 3.03 
            & 2.21$\times$ & 2.84
            & 2.29$\times$ & 2.82 \\

            & \mbox{DFlash~\tabsmall\secondarycolor{(16)}}
            & 5.21$\times$ & 6.59 
            & 6.18$\times$ & 7.87 
            & 5.67$\times$ & 7.13 
            & 5.21$\times$ & 6.52 
            & 4.71$\times$ & 5.98 
            & \textbf{5.37$\times$} & \textbf{7.12} 
            & 2.73$\times$ & 4.18 
            & 2.23$\times$ & 3.10
            & 4.66$\times$ & 6.06 \\

            & \mbox{\method~\tabsmall\secondarycolor{(16)}}
            & \textbf{7.92$\times$} & \textbf{10.03} 
            & \textbf{7.38$\times$} & \textbf{9.43} 
            & \textbf{5.85$\times$} & \textbf{7.41} 
            & \textbf{5.89$\times$} & \textbf{7.39} 
            & \textbf{5.53$\times$} & \textbf{7.04} 
            & 5.27$\times$ & 7.04 
            & \textbf{3.29$\times$} & \textbf{5.18} 
            & \textbf{2.78$\times$} & \textbf{3.87}
            & \textbf{5.49$\times$} & \textbf{7.17} \\

            \midrule
            \multicolumn{2}{c}{Temperature = 1} 
            & \tabsmall Speedup & \tabsmall $\tau$ 
            & \tabsmall Speedup & \tabsmall $\tau$ 
            & \tabsmall Speedup & \tabsmall $\tau$ 
            & \tabsmall Speedup & \tabsmall $\tau$ 
            & \tabsmall Speedup & \tabsmall $\tau$ 
            & \tabsmall Speedup & \tabsmall $\tau$ 
            & \tabsmall Speedup & \tabsmall $\tau$
            & \tabsmall Speedup & \tabsmall $\tau$
            & \tabsmall Speedup & \tabsmall $\tau$  \\
            
            \midrule
            \multirow{5}{*}{Qwen3-4B}
            & \mbox{EAGLE-3~\tabsmall\secondarycolor{(16)}}
            & 2.18$\times$ & 3.26 
            & 2.00$\times$ & 3.00 
            & 1.86$\times$ & 2.80 
            & 2.02$\times$ & 3.04 
            & 1.97$\times$ & 2.95 
            & 1.87$\times$ & 2.83 
            & 1.90$\times$ & 2.91 
            & 1.78$\times$ & 2.72
            & 1.95$\times$ & 2.94 \\

            & \mbox{EAGLE-3~\tabsmall\secondarycolor{(60)}} 
            & 2.44$\times$ & 3.77 
            & 2.28$\times$ & 3.51 
            & 2.08$\times$ & 3.10 
            & 2.09$\times$ & 3.09 
            & 2.19$\times$ & 3.38 
            & 2.05$\times$ & 3.19 
            & 2.11$\times$ & 3.33 
            & 2.02$\times$ & 3.17
            & 2.16$\times$ & 3.32 \\

            & \mbox{DART~\tabsmall\secondarycolor{(60)}}
            & 2.19$\times$ & 2.65 
            & 2.20$\times$ & 2.63 
            & 2.13$\times$ & 2.59 
            & 2.48$\times$ & 2.99 
            & 2.46$\times$ & 3.01 
            & 2.23$\times$ & 2.76 
            & 2.19$\times$ & 2.69 
            & 2.16$\times$ & 2.78
            & 2.26$\times$ & 2.76 \\

            & \mbox{DFlash~\tabsmall\secondarycolor{(16)}}
            & 4.66$\times$ & 5.98 
            & 5.02$\times$ & 6.58 
            & \textbf{3.79$\times$} & \textbf{5.05} 
            & 4.73$\times$ & 5.96 
            & 4.44$\times$ & 5.67 
            & 4.76$\times$ & 6.38 
            & 2.62$\times$ & 3.97 
            & 2.18$\times$ & 3.03
            & 4.03$\times$ & 5.33 \\

            & \mbox{\method~\tabsmall\secondarycolor{(16)}}
            & \textbf{6.79$\times$} & \textbf{8.61} 
            & \textbf{5.67$\times$} & \textbf{7.35} 
            & 3.75$\times$ & 4.83 
            & \textbf{5.02$\times$} & \textbf{6.30} 
            & \textbf{4.96$\times$} & \textbf{6.36} 
            & \textbf{5.04$\times$} & \textbf{6.45} 
            & \textbf{3.02$\times$} & \textbf{4.56} 
            & \textbf{2.59$\times$} & \textbf{3.52}
            & \textbf{4.61$\times$} & \textbf{6.00} \\

            \midrule
            \multirow{5}{*}{Qwen3-8B}
            & \mbox{EAGLE-3~\tabsmall\secondarycolor{(16)}}
            & 2.18$\times$ & 3.26 
            & 1.96$\times$ & 3.00 
            & 1.86$\times$ & 2.80 
            & 2.05$\times$ & 3.09 
            & 1.97$\times$ & 2.95 
            & 1.90$\times$ & 2.93 
            & 1.91$\times$ & 2.91 
            & 1.78$\times$ & 2.83
            & 1.95$\times$ & 2.97 \\

            & \mbox{EAGLE-3~\tabsmall\secondarycolor{(60)}} 
            & 2.40$\times$ & 3.70 
            & 2.23$\times$ & 3.44 
            & 2.05$\times$ & 3.17 
            & 2.32$\times$ & 3.58 
            & 2.11$\times$ & 3.24 
            & 1.90$\times$ & 2.95 
            & 1.91$\times$ & 3.02 
            & 1.78$\times$ & 2.81
            & 2.09$\times$ & 3.24 \\

            & \mbox{DART~\tabsmall\secondarycolor{(60)}}
            & 2.25$\times$ & 2.71 
            & 2.25$\times$ & 2.70 
            & 2.13$\times$ & 2.59 
            & 2.45$\times$ & 2.95 
            & 2.45$\times$ & 2.98 
            & 2.25$\times$ & 2.78 
            & 2.24$\times$ & 2.73 
            & 2.18$\times$ & 2.84
            & 2.28$\times$ & 2.79 \\

            & \mbox{DFlash~\tabsmall\secondarycolor{(16)}}
            & 4.71$\times$ & 6.00 
            & 5.00$\times$ & 6.50 
            & 3.72$\times$ & 4.77 
            & 4.34$\times$ & 5.44 
            & 4.14$\times$ & 5.23 
            & 5.04$\times$ & \textbf{6.73} 
            & 2.55$\times$ & 3.87 
            & 2.16$\times$ & 2.93
            & 3.96$\times$ & 5.18 \\

            & \mbox{\method~\tabsmall\secondarycolor{(16)}}
            & \textbf{6.47$\times$} & \textbf{8.34} 
            & \textbf{5.40$\times$} & \textbf{7.20} 
            & \textbf{3.78$\times$} & \textbf{4.92} 
            & \textbf{4.75$\times$} & \textbf{5.98} 
            & \textbf{4.73$\times$} & \textbf{6.02} 
            & \textbf{5.06$\times$} & 6.72 
            & \textbf{2.94$\times$} & \textbf{4.61} 
            & \textbf{2.52$\times$} & \textbf{3.46}
            & \textbf{4.46$\times$} & \textbf{5.91} \\

            \bottomrule
        \end{tabular}
    }
\end{table*}
\section{Experiments}

\label{sec:experiments}
\subsection{Experimental Setup}

\paragraph{Models and Evaluations.}
We evaluate Domino on Qwen3-4B, and Qwen3-8B \citep{qwen2025qwen3}. Following the evaluation protocol of DFlash \citep{chen2026dflash}, we consider tasks from three categories: math reasoning, code generation, and open-ended dialogue. For math reasoning, we evaluate on GSM8K \citep{cobbe2021training}, MATH \citep{hendrycks2021measuring}, and AIME25 \citep{maa2025aime}; for code generation, we use HumanEval \citep{chen2021evaluating}, MBPP \citep{austin2021program}, and LiveCodeBench \citep{jain2024livecodebench}; for dialogue, we evaluate on MT-Bench \citep{zheng2023judging} and Alpaca \citep{taori2023alpaca}. For each task, we report the average acceptance length \(\tau\) and the end-to-end decoding speedup over the autoregressive baseline.

\paragraph{Training Data.}
We train the draft modules on \texttt{mlabonne/open-perfectblend}\footnote{\url{https://huggingface.co/datasets/mlabonne/open-perfectblend}}, a instruction-tuning dataset with 1.42M samples covering chat, math, code, and general instruction-following tasks. We regenerate all responses using the corresponding target model rather than using the original dataset responses. Additional training details are provided in Appendix~\ref{app:training_details}.

\paragraph{Baselines.}
We compare Domino with vanilla autoregressive decoding and representative speculative decoding baselines, including EAGLE-3 \citep{li2025eagle3}, DFlash \citep{chen2026dflash}, DART \citep{liu2026dart}, and FR-Spec \citep{zhao2025fr}. EAGLE-3 represents autoregressive drafting; DFlash and DART represent parallel drafting methods that reduce repeated per-token draft computation; and FR-Spec represents vocabulary-efficient speculative decoding by reducing full-vocabulary LM-head projection cost.  

\paragraph{Implementation Details.}
For Domino, we use a draft block size of 16 for all target models and a 5-layer parallel draft backbone. The hidden dimension of the GRU causal encoder is set to 1024, and the hidden dimension of the low-rank correction head is set to 256. Unless otherwise specified, all experiments are conducted on NVIDIA A100-SXM4-80GB GPUs. 

\subsection{Main Results}

\paragraph{Low-concurrency case.}
Table~\ref{tab:main-results} reports the end-to-end speedup and average acceptance length on Qwen3 models under the Transformers backend. Domino consistently outperforms autoregressive drafting methods such as EAGLE-3, as well as parallel drafting baselines including DART and DFlash. Compared with DFlash, Domino further improves the average speedup from \(4.70\times\) to \(5.47\times\) on Qwen3-4B and from \(4.66\times\) to \(5.49\times\) on Qwen3-8B under greedy decoding (\(T=0\)). Similar gains hold under sampling decoding (\(T=1\)), where the average speedup increases from \(4.03\times\) to \(4.61\times\) on Qwen3-4B and from \(3.96\times\) to \(4.46\times\) on Qwen3-8B. These results show that causal correction improves draft quality with little additional overhead, leading to a higher end-to-end speedup.
\paragraph{High-concurrency case.}
We further evaluate serving throughput under different concurrency levels using SGLang. As shown in Table~\ref{tab:high-concurrency-tps}, Domino achieves higher throughput than EAGLE-3 and DFlash on both Qwen3-4B and Qwen3-8B. The gains indicate that the improved draft quality of Domino can be effectively translated into practical serving throughput, while maintaining the low-overhead advantage of block-parallel drafting.

\subsection{Ablation}
\label{ablation}
We conduct ablation studies to understand the source of Domino's improvement. 
Specifically, we examine whether the gain comes from differences in training data, the proposed training strategy, and the lightweight Domino head. 
Unless otherwise specified, all ablation experiments are conducted on Qwen3-8B with greedy decoding. 
The draft models are trained on ShareGPT, and evaluations are performed on NVIDIA A100 GPUs.
\subsubsection{Training Data}
\begin{table}[t]
\centering
\caption{High-concurrency throughput on Qwen3 models. Baseline rows report absolute throughput in TPS. Other entries report TPS with green subscripts indicating speedup over the corresponding baseline.}
\label{tab:high-concurrency-tps}
\resizebox{\columnwidth}{!}{
\scriptsize
\setlength{\tabcolsep}{2.0pt}
\renewcommand{\arraystretch}{0.92}
\begin{tabular}{l l r@{}l r@{}l r@{}l r@{}l r@{}l}
\toprule
\multirow{2}{*}{Task}
& \multirow{2}{*}{Method}
& \multicolumn{10}{c}{Concurrency} \\
\cmidrule(lr){3-12}
& & 2 & & 4 & & 8 & & 16 & & 32 & \\
\midrule

\multicolumn{12}{l}{\textbf{Qwen3-4B}} \\
\midrule

\multirow{5}{*}{GSM8K}
& Baseline
& 293 & & 557 & & 1079 & & 1884 & & 2868 & \\
& EAGLE-3 (16)
& 453 & \textcolor{green!45!black}{\scriptsize\ensuremath{_{1.5\times}}}
& 832 & \textcolor{green!45!black}{\scriptsize\ensuremath{_{1.5\times}}}
& 1375 & \textcolor{green!45!black}{\scriptsize\ensuremath{_{1.3\times}}}
& 1839 & \textcolor{green!45!black}{\scriptsize\ensuremath{_{1.0\times}}}
& 2170 & \textcolor{green!45!black}{\scriptsize\ensuremath{_{0.8\times}}} \\
& EAGLE-3 (60)
& 458 & \textcolor{green!45!black}{\scriptsize\ensuremath{_{1.6\times}}}
& 683 & \textcolor{green!45!black}{\scriptsize\ensuremath{_{1.2\times}}}
& 896 & \textcolor{green!45!black}{\scriptsize\ensuremath{_{0.8\times}}}
& 1040 & \textcolor{green!45!black}{\scriptsize\ensuremath{_{0.6\times}}}
& 1133 & \textcolor{green!45!black}{\scriptsize\ensuremath{_{0.4\times}}} \\
& DFlash (16)
& 965 & \textcolor{green!45!black}{\scriptsize\ensuremath{_{3.3\times}}}
& 1698 & \textcolor{green!45!black}{\scriptsize\ensuremath{_{3.0\times}}}
& 2738 & \textcolor{green!45!black}{\scriptsize\ensuremath{_{2.5\times}}}
& 3538 & \textcolor{green!45!black}{\scriptsize\ensuremath{_{1.9\times}}}
& 4397 & \textcolor{green!45!black}{\scriptsize\ensuremath{_{1.5\times}}} \\
& \textbf{\textsc{Domino} (16)}
& \textbf{1256} & \textcolor{green!45!black}{\scriptsize\ensuremath{_{4.3\times}}}
& \textbf{2202} & \textcolor{green!45!black}{\scriptsize\ensuremath{_{4.0\times}}}
& \textbf{3441} & \textcolor{green!45!black}{\scriptsize\ensuremath{_{3.2\times}}}
& \textbf{4467} & \textcolor{green!45!black}{\scriptsize\ensuremath{_{2.4\times}}}
& \textbf{5509} & \textcolor{green!45!black}{\scriptsize\ensuremath{_{1.9\times}}} \\

\cmidrule(lr){1-12}

\multirow{5}{*}{MBPP}
& Baseline
& 291 & & 544 & & 1002 & & 1586 & & 2300 & \\
& EAGLE-3 (16)
& 407 & \textcolor{green!45!black}{\scriptsize\ensuremath{_{1.4\times}}}
& 740 & \textcolor{green!45!black}{\scriptsize\ensuremath{_{1.4\times}}}
& 1186 & \textcolor{green!45!black}{\scriptsize\ensuremath{_{1.2\times}}}
& 1597 & \textcolor{green!45!black}{\scriptsize\ensuremath{_{1.0\times}}}
& 1892 & \textcolor{green!45!black}{\scriptsize\ensuremath{_{0.8\times}}} \\
& EAGLE-3 (60)
& 414 & \textcolor{green!45!black}{\scriptsize\ensuremath{_{1.4\times}}}
& 610 & \textcolor{green!45!black}{\scriptsize\ensuremath{_{1.1\times}}}
& 793 & \textcolor{green!45!black}{\scriptsize\ensuremath{_{0.8\times}}}
& 899 & \textcolor{green!45!black}{\scriptsize\ensuremath{_{0.6\times}}}
& 987 & \textcolor{green!45!black}{\scriptsize\ensuremath{_{0.4\times}}} \\
& DFlash (16)
& 914 & \textcolor{green!45!black}{\scriptsize\ensuremath{_{3.1\times}}}
& 1650 & \textcolor{green!45!black}{\scriptsize\ensuremath{_{3.0\times}}}
& 2501 & \textcolor{green!45!black}{\scriptsize\ensuremath{_{2.5\times}}}
& 3330 & \textcolor{green!45!black}{\scriptsize\ensuremath{_{2.1\times}}}
& 4088 & \textcolor{green!45!black}{\scriptsize\ensuremath{_{1.8\times}}} \\
& \textbf{\textsc{Domino} (16)}
& \textbf{968} & \textcolor{green!45!black}{\scriptsize\ensuremath{_{3.3\times}}}
& \textbf{1654} & \textcolor{green!45!black}{\scriptsize\ensuremath{_{3.0\times}}}
& \textbf{2651} & \textcolor{green!45!black}{\scriptsize\ensuremath{_{2.6\times}}}
& \textbf{3422} & \textcolor{green!45!black}{\scriptsize\ensuremath{_{2.2\times}}}
& \textbf{4290} & \textcolor{green!45!black}{\scriptsize\ensuremath{_{1.9\times}}} \\

\midrule
\multicolumn{12}{l}{\textbf{Qwen3-8B}} \\
\midrule

\multirow{5}{*}{GSM8K}
& Baseline
& 184 & & 360 & & 655 & & 1143 & & 1713 & \\
& EAGLE-3 (16)
& 324 & \textcolor{green!45!black}{\scriptsize\ensuremath{_{1.8\times}}}
& 598 & \textcolor{green!45!black}{\scriptsize\ensuremath{_{1.7\times}}}
& 979 & \textcolor{green!45!black}{\scriptsize\ensuremath{_{1.5\times}}}
& 1276 & \textcolor{green!45!black}{\scriptsize\ensuremath{_{1.1\times}}}
& 1398 & \textcolor{green!45!black}{\scriptsize\ensuremath{_{0.8\times}}} \\
& EAGLE-3 (60)
& 330 & \textcolor{green!45!black}{\scriptsize\ensuremath{_{1.8\times}}}
& 482 & \textcolor{green!45!black}{\scriptsize\ensuremath{_{1.3\times}}}
& 577 & \textcolor{green!45!black}{\scriptsize\ensuremath{_{0.9\times}}}
& 624 & \textcolor{green!45!black}{\scriptsize\ensuremath{_{0.5\times}}}
& 699 & \textcolor{green!45!black}{\scriptsize\ensuremath{_{0.4\times}}} \\
& DFlash (16)
& 672 & \textcolor{green!45!black}{\scriptsize\ensuremath{_{3.7\times}}}
& 1243 & \textcolor{green!45!black}{\scriptsize\ensuremath{_{3.4\times}}}
& 1915 & \textcolor{green!45!black}{\scriptsize\ensuremath{_{2.9\times}}}
& 2533 & \textcolor{green!45!black}{\scriptsize\ensuremath{_{2.2\times}}}
& 2801 & \textcolor{green!45!black}{\scriptsize\ensuremath{_{1.6\times}}} \\
& \textbf{\textsc{Domino} (16)}
& \textbf{942} & \textcolor{green!45!black}{\scriptsize\ensuremath{_{5.1\times}}}
& \textbf{1703} & \textcolor{green!45!black}{\scriptsize\ensuremath{_{4.7\times}}}
& \textbf{2678} & \textcolor{green!45!black}{\scriptsize\ensuremath{_{4.1\times}}}
& \textbf{3379} & \textcolor{green!45!black}{\scriptsize\ensuremath{_{3.0\times}}}
& \textbf{3650} & \textcolor{green!45!black}{\scriptsize\ensuremath{_{2.1\times}}} \\

\cmidrule(lr){1-12}

\multirow{5}{*}{MBPP}
& Baseline
& 183 & & 352 & & 635 & & 1076 & & 1428 & \\
& EAGLE-3 (16)
& 290 & \textcolor{green!45!black}{\scriptsize\ensuremath{_{1.6\times}}}
& 525 & \textcolor{green!45!black}{\scriptsize\ensuremath{_{1.5\times}}}
& 886 & \textcolor{green!45!black}{\scriptsize\ensuremath{_{1.4\times}}}
& 1180 & \textcolor{green!45!black}{\scriptsize\ensuremath{_{1.1\times}}}
& 1291 & \textcolor{green!45!black}{\scriptsize\ensuremath{_{0.9\times}}} \\
& EAGLE-3 (60)
& 293 & \textcolor{green!45!black}{\scriptsize\ensuremath{_{1.6\times}}}
& 438 & \textcolor{green!45!black}{\scriptsize\ensuremath{_{1.2\times}}}
& 525 & \textcolor{green!45!black}{\scriptsize\ensuremath{_{0.8\times}}}
& 574 & \textcolor{green!45!black}{\scriptsize\ensuremath{_{0.5\times}}}
& 631 & \textcolor{green!45!black}{\scriptsize\ensuremath{_{0.4\times}}} \\
& DFlash (16)
& 649 & \textcolor{green!45!black}{\scriptsize\ensuremath{_{3.6\times}}}
& 1169 & \textcolor{green!45!black}{\scriptsize\ensuremath{_{3.3\times}}}
& 1889 & \textcolor{green!45!black}{\scriptsize\ensuremath{_{3.0\times}}}
& 2487 & \textcolor{green!45!black}{\scriptsize\ensuremath{_{2.3\times}}}
& 2800 & \textcolor{green!45!black}{\scriptsize\ensuremath{_{2.0\times}}} \\
& \textbf{\textsc{Domino} (16)}
& \textbf{701} & \textcolor{green!45!black}{\scriptsize\ensuremath{_{3.8\times}}}
& \textbf{1256} & \textcolor{green!45!black}{\scriptsize\ensuremath{_{3.6\times}}}
& \textbf{2035} & \textcolor{green!45!black}{\scriptsize\ensuremath{_{3.2\times}}}
& \textbf{2727} & \textcolor{green!45!black}{\scriptsize\ensuremath{_{2.5\times}}}
& \textbf{3027} & \textcolor{green!45!black}{\scriptsize\ensuremath{_{2.1\times}}} \\

\bottomrule
\end{tabular}
}
\vspace{-2mm}
\end{table}
To isolate the impact of model architecture, we train all baselines on identical data. During evaluation, we apply greedy decoding with a fixed 16-token drafting budget across all methods.

Table~\ref{tab:same-data-abl} shows a clear trade-off between acceptance length and drafting overhead. EAGLE-3 achieves strong acceptance lengths, e.g., \(5.01\) on GSM8K, but its throughput is limited by sequential drafting. DFlash has lower acceptance lengths, e.g., \(3.90\) on GSM8K and \(3.78\) on HumanEval, but obtains higher throughput by using parallel drafting. Domino achieves a better balance: it reaches comparable acceptance lengths to autoregressive methods while maintaining low drafting overhead, resulting in the best throughput on almost all tasks and concurrency levels. This confirms that the gain comes from the proposed design rather than training-data differences.

\subsubsection{Training Strategy}
\label{sec: abl_training}
\begin{figure}[t]
    
    \centering
    \includegraphics[width=\columnwidth]{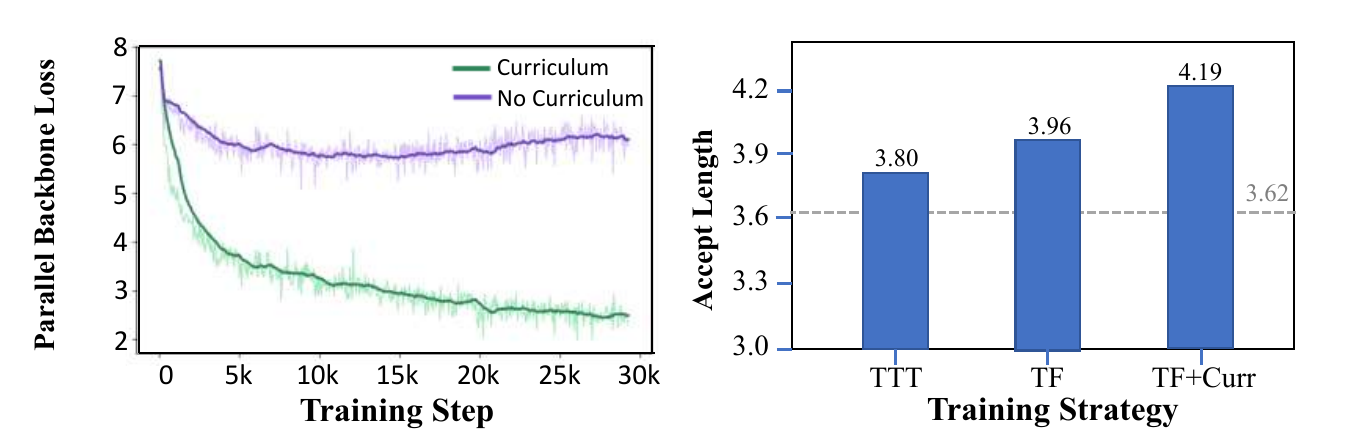}
    \caption{
    Left: parallel backbone loss with and without the base-anchored curriculum. 
    Right: average acceptance length under TTT, TF, and TF+Curr. 
    TTT denotes training-time testing, TF denotes teacher forcing, and Curr denotes the base-anchored curriculum. 
    The gray dashed line denotes the DFlash reference.
    }
    \label{fig:training_strategy}
    \vspace{-2mm}
\end{figure}

We ablate the training strategy for the causal correction branch in Figure~\ref{fig:training_strategy}. Compared with training-time test (TTT), teacher forcing improves the average acceptance length from \(3.80\) to \(3.96\). This supports our motivation that the causal encoder should be trained on ground-truth prefixes rather than noisy self-generated prefixes, since only draft positions whose previous tokens have been accepted can contribute to the final acceptance length.

However, direct teacher forcing alone is not sufficient. As discussed in Section~\ref{sec:training}, clean ground-truth prefixes can allow the correction branch to shortcut the parallel backbone, causing the base logits to collapse. This is reflected in Figure~\ref{fig:training_strategy}, where direct final-logit training keeps the parallel backbone loss high. The base-anchored curriculum mitigates this issue by first strengthening the base logits and then gradually shifting optimization toward the final logits. As a result, TF+Curriculum further improves the average acceptance length from \(3.96\) to \(4.19\). These results validate both parts of our training strategy: teacher forcing provides a more useful learning signal for causal correction, while the base-anchored curriculum prevents backbone collapse and improves the final draft quality.
\begin{table}[t]
\caption{Same-data comparison on Qwen3-8B under greedy decoding. All draft models are trained on ShareGPT with the same 16-token drafting budget. FR-Spec uses a 32K frequency-ranked vocabulary subset from ShareGPT. Baseline reports TPS; other methods report speedup over the baseline.}
\label{tab:same-data-abl}
\centering
\setlength{\tabcolsep}{4pt}
\renewcommand{\arraystretch}{0.95}

\resizebox{\linewidth}{!}{
\begin{tabular}{l cccccc c}
\toprule
\multirow{2}[2]{*}{Method} 
& \multicolumn{6}{c}{Concurrency} 
& \multirow{2}[2]{*}{\textit{Avg.} $\tau$} \\
\cmidrule(lr){2-7}
& 1 & 2 & 4 & 8 & 16 & 32 &  \\
\midrule

\multicolumn{8}{l}{\textbf{GSM8K}} \\
\midrule
Baseline (TPS)
& 92 & 184 & 360 & 655 & 1143 & 1713 & -- \\

EAGLE-3
& 2.35$\times$ & 2.15$\times$ & 1.90$\times$ & 1.77$\times$ & 1.30$\times$ & 0.97$\times$
& 5.01 \\

FR-Spec
& 2.77$\times$ & 2.52$\times$ & 2.36$\times$ & 2.09$\times$ & 1.53$\times$ & 1.16$\times$
& 4.79 \\

DFlash
& 2.68$\times$ & 2.40$\times$ & 2.30$\times$ & 1.95$\times$ & 1.49$\times$ & 1.09$\times$
& 3.90 \\

\textbf{\textsc{Domino}}
& \textbf{3.01$\times$} & \textbf{2.70$\times$} & \textbf{2.61$\times$} & \textbf{2.18$\times$} & \textbf{1.68$\times$} & \textbf{1.24$\times$}
& 4.65 \\

\midrule
\multicolumn{8}{l}{\textbf{HumanEval}} \\
\midrule
Baseline (TPS)
& 92 & 183 & 355 & 667 & 1211 & 2036 & -- \\

EAGLE-3
& 2.27$\times$ & 2.13$\times$ & 2.00$\times$ & 1.74$\times$ & 1.27$\times$ & 0.87$\times$
& 4.84 \\

FR-Spec
& 2.67$\times$ & 2.43$\times$ & 2.30$\times$ & 2.03$\times$ & 1.50$\times$ & 1.02$\times$
& 4.54 \\

DFlash
& 2.58$\times$ & 2.43$\times$ & 2.32$\times$ & 2.05$\times$ & 1.47$\times$ & 1.00$\times$
& 3.78 \\

\textbf{\textsc{Domino}}
& \textbf{2.82$\times$} & \textbf{2.64$\times$} & \textbf{2.52$\times$} & \textbf{2.23$\times$} & \textbf{1.63$\times$} & \textbf{1.12$\times$}
& 4.35 \\

\midrule
\multicolumn{8}{l}{\textbf{LiveCodeBench}} \\
\midrule
Baseline (TPS)
& 91 & 175 & 283 & 569 & 744 & 1140 & -- \\

EAGLE-3
& 1.99$\times$ & 1.84$\times$ & 1.81$\times$ & 1.41$\times$ & 1.30$\times$ & 0.92$\times$
& 4.38 \\

FR-Spec
& 2.36$\times$ & 2.18$\times$ & 2.17$\times$ & 1.64$\times$ & 1.51$\times$ & 1.10$\times$
& 4.21 \\

DFlash
& 2.36$\times$ & 2.15$\times$ & 2.16$\times$ & 1.56$\times$ & 1.56$\times$ & 1.10$\times$
& 3.66 \\

\textbf{\textsc{Domino}}
& \textbf{2.55$\times$} & \textbf{2.33$\times$} & \textbf{2.44$\times$} & \textbf{1.75$\times$} & \textbf{1.60$\times$} & \textbf{1.17$\times$}
& 4.24 \\

\bottomrule
\end{tabular}
}
\end{table}

\subsubsection{Effect of Domino Head}

We ablate the Domino head by evaluating the same model with the causal correction branch disabled or enabled. As shown in Table~\ref{tab:domino_head_avg}, enabling the Domino head improves the average acceptance length from \(3.49\) to \(4.19\) and the average speedup from \(2.84\times\) to \(3.31\times\). This confirms that lightweight prefix-dependent correction is the key source of the improvement over the parallel backbone alone.

\begin{table}[t]
\centering
\caption{
Effect of the Domino head. We report average results here and provide full results in Table~\ref{tab:domino_head_full}.
}
\label{tab:domino_head_avg}
\setlength{\tabcolsep}{5pt}
\renewcommand{\arraystretch}{1.05}
\resizebox{\columnwidth}{!}{
\begin{tabular}{lcc}
\toprule
Method & Avg. Accept Length & Avg. Speedup \\
\midrule
w/o Domino Head & 3.49 & 2.84$\times$ \\
w/ Domino Head  & \textbf{4.19} & \textbf{3.31}$\times$ \\
\bottomrule
\end{tabular}
}
\vspace{-2mm}
\end{table}

\section{Conclusion}
We propose Domino, a speculative decoding framework that improves block-parallel drafting with lightweight causal correction. Domino decouples causal dependency from expensive autoregressive execution, improving draft quality while maintaining low drafting overhead. Experiments on Qwen3 models demonstrate consistent gains in acceptance length and end-to-end speedup over representative baselines. These results suggest that causal information can be effectively reintroduced into parallel drafting for faster LLM inference.

\section{Limitations}
This work focuses on inference acceleration rather than reducing the cost of model training or finetuning.
Our current implementation is mainly adapted to SGLang, and its compatibility with other serving frameworks remains to be systematically evaluated.
Moreover, the practical speedup can vary across hardware platforms due to differences in memory bandwidth, compute capability, and kernel efficiency.
As a result, further platform-specific optimization may be needed for deployment in different environments.

\newpage
\bibliography{custom}

\appendix

\section{Appendix}
\label{sec:appendix}
\subsection{Training Details}
\label{app:training_details}

For both Qwen3-4B and Qwen3-8B, we train the \method draft module while keeping the target model frozen. We use the regenerated PerfectBlend data described in Section~\ref{sec:experiments}. Input sequences are truncated to a maximum length of 3072 tokens, and the draft block size is set to 16.

Unless otherwise specified, all draft modules are trained for 3 epochs on 8 NVIDIA A100-SXM4-80GB GPUs. We use a per-GPU batch size of 2, resulting in a global batch size of 16 without gradient accumulation. We optimize the model with AdamW using a learning rate of \(6\times 10^{-4}\), zero weight decay, gradient clipping with a maximum norm of 1.0, and a cosine learning-rate schedule with a warmup ratio of 0.04. Training is conducted in bfloat16 precision with FSDP and gradient sharding.

\begin{table}[H]
\centering
\caption{Baseline draft model checkpoints used in our experiments.}
\label{tab:baseline_checkpoints}
\resizebox{\linewidth}{!}{
\begin{tabular}{lcc}
\toprule
\textbf{Method} & \textbf{Qwen3-4B} & \textbf{Qwen3-8B} \\
\midrule
EAGLE-3 
& \texttt{Angslim/Qwen3-4B\_eagle3} 
& \texttt{Angslim/Qwen3-8B\_eagle3} \\

DART 
& \texttt{fvliang/qwen4b-dart} 
& \texttt{fvliang/qwen8b-dart} \\

DFlash 
& \texttt{z-lab/Qwen3-4B-DFlash-b16} 
& \texttt{z-lab/Qwen3-8B-DFlash-b16} \\
\bottomrule
\end{tabular}
}
\end{table}

\begin{table}[H]
\centering
\caption{
Full benchmark-level results for the Domino head ablation. The same trained model is evaluated with the causal correction branch disabled or enabled.
}
\label{tab:domino_head_full}
\setlength{\tabcolsep}{4pt}
\renewcommand{\arraystretch}{1.05}
\resizebox{\columnwidth}{!}{
\begin{tabular}{lcccc}
\toprule
\multirow{2}{*}{Benchmark} 
& \multicolumn{2}{c}{w/o Domino Head} 
& \multicolumn{2}{c}{w/ Domino Head} \\
\cmidrule(lr){2-3}
\cmidrule(lr){4-5}
& Accept Length & Speedup 
& Accept Length & Speedup \\
\midrule
GSM8K         & 3.82 & 3.17$\times$ & \textbf{4.80} & \textbf{3.84}$\times$ \\
MATH-500      & 3.76 & 3.08$\times$ & \textbf{4.66} & \textbf{3.74}$\times$ \\
AIME25        & 3.68 & 3.03$\times$ & \textbf{4.29} & \textbf{3.47}$\times$ \\
HumanEval     & 3.69 & 3.04$\times$ & \textbf{4.35} & \textbf{3.51}$\times$ \\
MBPP          & 3.30 & 2.70$\times$ & \textbf{3.94} & \textbf{3.16}$\times$ \\
LiveCodeBench & 3.36 & 2.64$\times$ & \textbf{3.92} & \textbf{2.98}$\times$ \\
MT-Bench      & 2.84 & 2.21$\times$ & \textbf{3.39} & \textbf{2.46}$\times$ \\
\midrule
Avg.          & 3.49 & 2.84$\times$ & \textbf{4.19} & \textbf{3.31}$\times$ \\
\bottomrule
\end{tabular}
}
\vspace{-2mm}
\end{table}

\end{document}